\def\year{2021}\relax

\documentclass[conference]{IEEEtran}
\IEEEoverridecommandlockouts
\usepackage{times}  %
\usepackage{helvet} %
\usepackage{courier}  %
\usepackage[hyphens]{url}  %
\usepackage{graphicx}
\usepackage{booktabs}
\usepackage{mathtools}
\usepackage{amsmath}
\usepackage{algorithm,algorithmicx,algpseudocode}
\usepackage{multirow}
\usepackage[export]{adjustbox}
\usepackage[table]{xcolor}
\usepackage{array}
\usepackage{setspace}
\usepackage{cite}
 
\usepackage{times}  %
\usepackage{helvet} %
\usepackage{courier}  %
\usepackage[hyphens]{url}  %
\usepackage{caption} %
\usepackage{amssymb}
\usepackage{subcaption}

\usepackage[switch]{lineno}
\usepackage{algorithm,algorithmicx,algpseudocode}
\usepackage{multirow}
\usepackage[export]{adjustbox}
\usepackage[table]{xcolor}
\usepackage{array}
\usepackage{setspace}
\usepackage{commath}
\usepackage{comment}
\newcolumntype{C}[1]{>{\centering\let\newline\\\arraybackslash\hspace{0pt}}m{#1}}

\title{Adversarial Robustness via Stochastic Regularization \\ of Neural Activation Sensitivity}

\author{Gil Fidel\qquad Ron Bitton\qquad Ziv Katzir\qquad Asaf Shabtai \\ 
Department of Software and Information Systems Engineering \\ 
Ben-Gurion university of the Negev \\
\{fidelg,ronbit,zivka\}@post.bgu.ac.il, shabtaia@bgu.ac.il}

\begin{document}

\maketitle

\begin{abstract}

Recent works have shown that the input domain of any machine learning classifier is bound to contain adversarial examples. 
Thus, we can no longer hope to immune classifiers against adversarial examples and instead can only aim to achieve the following two defense goals: 1) make adversarial examples harder to find, or 2) weaken their adversarial nature by pushing them further away from correctly classified data points.
Most, if not all, of the previously suggested defense mechanisms address just one of those two goals and as such, can be bypassed by adaptive attacks that take the defense mechanism into consideration. 
In this work, we suggest a novel defense mechanism that simultaneously addresses both defense goals: 
We flatten the gradients of the loss surface, making adversarial examples harder to find by using a novel stochastic regularization term that explicitly decreases the sensitivity of individual neurons to small input perturbations.
In addition, we push the decision boundary away from correctly classified inputs by leveraging Jacobian regularization. 
We present a solid theoretical basis and an empirical testing of our suggested approach, demonstrate its superiority over previously suggested defense mechanisms, and show that it is effective against a wide range of adaptive attacks.

\end{abstract}

\section{Introduction}
Recently published works have established the inevitability of adversarial examples~\cite{shafahi2018adversarial, cohen2019certified, shamir2019simple, katzir2020gradients}.
Despite a few theoretical aspects that still require further research attention, these publications collectively imply that adversarial examples exist within the input space of each and every classification model. 
That is, the input space of any classifier is bound to include incorrectly classified input samples which are located in proximity to correctly classified input. 
The existence of such adversarial data points is unavoidable regardless of the learning algorithm, training dataset, or hyperparameters used in creating the model.
Accepting the inevitability of adversarial examples means that defenders can only hope to \textit{a)} make such examples harder to find, or \textit{b)} weaken their adversarial nature by reducing their similarity to correctly classified inputs. 
These two defense goals are largely independent of one another, and often a given defense mechanism addresses just one of them. 

Many of the commonly used methods for finding adversarial examples rely on the gradients of the classifier's loss surface~\cite{goodfellow2014explaining, kurakin2016adversarial, goodfellow2016dlbook, carlini2017towards}. 
Those methods utilize greedy, iterative optimization schemes (e.g., stochastic gradient descent) to traverse the loss surface in search of the smallest perturbation that will cause the classifier to predict an incorrect class label. 
For this reason, making adversarial examples harder to find is typically associated with \textit{gradient obfuscation}~\cite{papernot2017practical}.
That is, many of the suggested approaches for increasing the difficulty of finding adversarial examples attempt to either eliminate the loss gradient or alternatively to cause the behaviour of the gradient to be inconsistent in proximity to known training samples. 
The result, in both cases, is that gradient-based optimization becomes considerably harder to implement. 

More recently, efficient approaches for traversing the classifier decision boundary itself~\cite{chen2019hopskipjumpattack}, as opposed to the loss surface, were suggested. 
Those methods are able to produce highly refined adversarial examples without relying on the classifier's loss or its gradient. 
In fact, such methods are applicable even when the classifier's loss function is non-differentiable. 
In order to mitigate such attack methods, a defender must `push' the decision boundary away from the correctly classified training samples, increasing the \textit{mean perturbation distance} (i.e., the distance between a valid input and the corresponding adversarial examples), and in so doing, weaken the adversarial nature of the resulting adversarial examples.

A large number of defense mechanisms against adversarial perturbations have %
been proposed. 
Some attempted to increase the complexity of finding relevant adversarial examples~\cite{Xiao2020Enhancing,pang2019rethinking,Sen2020EMPIR}, while others aimed to increase the distance between correctly classified inputs and the classifier's decision boundary~\cite{hoffman2019robust,Jakubovitz_2018}.
However, most, if not all, of those approaches were quickly bypassed by tailor-made, \textit{adaptive attacks}~\cite{athalye2018obfuscated,carlini2017adversarial, tramer2020adaptive}; i.e., if an adversary is aware of the existence of a defense mechanism, he/she can adapt the attack method in order to bypass that defense. 
A defense mechanism that can address both defense goals listed above and sustain adaptive attacks has yet to be discovered.

In this work, we suggest a novel approach for increasing the resilience of neural network-based classifiers to adversarial attacks. 
Our approach aims to simultaneously increase the difficulty in finding adversarial examples and increase the mean perturbation distance achieved via black-box boundary-based approaches.
Increased complexity is achieved by introducing stochastic noise into the network's training phase and explicitly regularizing the changes inflicted to the activation values of individual neurons.
Pushing the decision boundary away from correctly classified inputs is done based on the well-studied Jacobian regularization~\cite{drucker1992improving} technique. 

Instead of attempting to completely deflect adversarial attacks, our goal is to render optimization-based attack methods ineffective, while at the same time increasing the mean perturbation distance so that the resulting perturbation can no longer be considered `similar' to the valid input. 
We  test our approach extensively using the popular MNIST~\cite{lecun1998gradient} and CIFAR-10~\cite{krizhevsky2009learning} benchmark datasets.
We follow current best practices~\cite{carlini2019evaluating} and test our suggested approach against generic gradient-based (white-box) attacks, tailor-made adaptive attacks, as well as a (black-box) boundary-based attack. 
Our results show that a training process which combines the two techniques results in a model that is robust to both gradient-based and boundary-based attacks.

\section{\label{sec:background}Background}
\subsection{Notations \& Preliminaries}
In the context of a  model $f\left(\cdot \right)$, an input vector ${x}'$ is considered to be an adversarial example if it is incorrectly classified by $f\left(\cdot \right)$ despite being located in proximity to some other, correctly classified, input $x$. More formally, given some distance metric $d$ and a small positive constant $\epsilon$, we say that ${x}'$ is a non-targeted adversarial example when $d\left(x, {x}'\right)<\epsilon$ and $f\left(x\right) \neq f\left({x}'\right)$. We refer to ${x}'$ as a targeted adversarial example if $f\left({x}'\right)$ equals some target class $t$ chosen by the adversary.
In the image classification domain, which we use in this work in order to evaluate our proposed method, three distance metrics are typically used: $L_{0}$ -- measures the number of perturbed input features (pixels), $L_{2}$ -- measures the Euclidean distance between the two input vectors, and $L_{\infty}$ -- measures the maximal change applied to any single feature.

\subsection{The Inevitability of Adversarial Examples}
The initial discovery of adversarial examples in deep neural networks (DNNs) unleashed an arms race of attack and defense methods.
However, for the most part, attackers seem to have the upper hand in this process. 
Numerous methods for creating classification models that are resilient to adversarial manipulations were suggested, however they were all %
bypassed by new, more sophisticated attacks. 
This has led researchers to wonder whether adversarial examples are in fact an inevitable artifact of DNNs or even of all machine learning (ML) based classifiers. 

A number of recent works have made considerable progress in proving the \textit{adversarial inevitability assumption}, making it safe to assume that the input spaces of all ML classifiers are bound to include adversarial examples~\cite{shafahi2018adversarial,shamir2019simple,gourdeau2019hardness,katzir2020gradients}.
\cite{gourdeau2019hardness} show that adversarial resilience cannot be achieved with a training dataset that is polynomial to the number of input dimensions, which means that adversarial examples are inevitable for all practical purposes.
\cite{katzir2020gradients} prove that in order to fully block adversarial attack methods, one should eliminate the network's ability to learn. 
Alternatively, any new defense mechanisms suggested should include a suitable training procedure that can balance the two contradicting needs.

\subsection{Identifying Adversarial Examples}%
Knowing that adversarial examples exist within the input space of the target classifier, the adversary does not actually create adversarial examples, but rather searches for them in the vicinity of the relevant input sample.
Therefore, it would be more accurate to discuss \textit{adversarial search methods} instead of \textit{adversarial attack methods}. 
We will %
use the two terms \textit{attack} and \textit{search} interchangeably throughout the rest of this paper.

Adversarial attack/search methods can largely be divided into two %
categories: methods that traverse the classifier's loss surface and methods that explore the classifier's decision boundary itself. 
Loss surface traversal approaches typically use gradient-based optimization methods, such as stochastic gradient descent.
Given some correctly classified input, those methods attempt to maximize classification loss while at the same time minimizing the perturbation distance. 
Early loss surface-based attacks leveraged simple heuristics with respect to the structure of the loss surface~\cite{goodfellow2014explaining, kurakin2016adversarial}.
This concept was later enhanced by Carlini and Wagner~\cite{carlini2017towards} who were the first to define a gradient-based optimization process in order to identify adversarial examples. 
The core principles presented in their work were later used in a large number of additional attack methods~\cite{chen2017zoo, chen2019hopskipjumpattack, sharif2016accessorize}.

Decision boundary-based attack methods operate by traversing the decision boundary itself.
They are unaffected by the structure of the loss surface and can be used even if the classifier itself is based on some non-differentiable element.
Such models define boundary surfaces between different classes (whether explicitly or implicitly), which can then be traversed in order to find a minimal perturbation given some source point. 
A notable example of this class of attack methods is HopSkipJump~\cite{chen2019hopskipjumpattack}, which has shown to be applicable to all ML-based classification models.
This method starts by processing two input points together: a source point and an additional data point which belongs to the desired target class. 
It uses binary search in order to find the boundary point that resides on the line connecting the two input points; then it approximates the gradient of the boundary surface at the boundary point using Monte Carlo search and uses this approximation in order to gradually minimize the perturbation distance. 
Notably, the HopSkipJump approach is applicable to all ML-based classification models. 
Such models define boundary surfaces between different classes (whether explicitly or implicitly), which can then be traversed in order to find a minimal perturbation given some source point. 

\subsection{Defending Against Adversarial Perturbations}
In recent years, numerous defense mechanisms addressing the two attack categories mentioned above were suggested. 
However, most, if not all, of those defense methods were later bypassed using stronger, tailor-made adversarial search methods referred to as \textit{Adaptive Attacks}. 
To the best of our knowledge, none of the defense methods suggested so far have simultaneously addressed both loss surface and boundary-based searches. 
As a result, defense mechanisms that prevented loss surface-based attacks remained susceptible to boundary-based attacks and vice versa.

Defending against loss surface-based attacks is generally done by obfuscating or eliminating the loss gradient in the vicinity of correctly classified inputs. 
A promising line of research which has been explored by several groups is based on injecting stochastic noise into the inference process. 
Defenses based on this concept have empirically demonstrated improved robustness to adversarial examples~\cite{roth2019odds,pang2019mixup} and were relatively difficult to overcome. Recently, a method called randomized smoothing was suggested~\cite{cohen2019certified} in order to allow the classifier to move away from the adversarial data point into a correctly classified input subspace. 
This method randomly picks multiple data points in the vicinity of the original input, uses the classifier to infer a class label for each of those random points, as well as the original input, and eventually treats the majority vote as the correct class label.
In~\cite{gilmer2019adversarial}, the authors were able to support the empirical results by proving that adversarial resilience is correlated with the model's resilience to Gaussian noise.

Our proposed approach differs from earlier works in two key respects: \textit{a)} we inject stochastic noise into the network's training process, as opposed to other methods that rely on randomization during inference, and \textit{b)} our random noise is selected from a much wider neighborhood of the original input. 
As will be described in upcoming sections, we believe that those two properties make our proposed method considerably harder to overcome using adaptive attacks. 

Defending against decision boundary-based attacks is done by increasing the generalization margins, and in so doing, `pushing' the decision boundary away from correctly classified data points. 
A number of works have explored different variants of Jacobian regularization in order to achieve this goal ~\cite{hoffman2019robust, Jakubovitz_2018}. 

\subsection{Adaptive Defense Evaluation}
The term \textit{adaptive attacks} refers to  adversarial attack methods that were tailored to counter a specific defense mechanism, where the attacker has complete knowledge of the defense mechanism's implementation details. 
The success of adaptive methods in bypassing previously suggested defense mechanisms has turned adaptive evaluation into a mandatory step prior to the publication of a new defense mechanism. 
Using cryptanalysis related terminology, any defense method that fails to handle adaptive attacks is in essence a form of 'security by obscurity.' It is based on the defender's ability to keep some parts of the defense mechanism hidden from the attacker.

Adaptive attacks were first suggested in the context of loss surface-based attack methods~\cite{athalye2018obfuscated}. 
Most notably, this work suggests using the expectancy of the loss gradient measured over multiple inference acts of the same input in order to counter the effect of stochastic noise injected at inference time. 
Another related approach, which is commonly known as random restart, executes a base attack method while slightly perturbing the input point with random noise each time. 
As will be discussed later, both of the approaches are relevant when the stochastic element of the suggested defense mechanism is added during inference time. Our proposed defense mechanism applies randomization to the network's original training phase and by that dramatically reduces the effectiveness of the aforementioned adaptive methods.

Recently, the scope of adaptive attack methods was extended to include black-box boundary-based attacks as part of the `adaptive toolbox'~\cite{tramer2020adaptive}. 
As part of this work, the authors devise adaptive attacks against 13 recently published defense mechanisms. 
In two of the 13 cases, loss gradient-based approaches obtain poor results or simply require too much computational effort, leading the team to use black-box boundary attacks instead. 
This understanding supports our claim that any defense mechanism against adversarial examples must simultaneously address both loss surface and decision boundary-based attack methods. 

As described in upcoming sections of this work, we evaluate our proposed method against a range of adaptive attack methods and find it to be superior to previously suggested defense mechanisms.
We attribute this superiority to the combined nature of our approach.

\section{\label{sec:proposed_method}Proposed method}
\subsection{Method Overview}
The proposed method aims to simultaneously increase the difficulty in finding adversarial examples and push the decision boundary away from correctly classified data points.
To do so, the method integrates two independent regularization terms into the classifier network's loss function: 1) in order to increase the complexity of finding adversarial examples, we introduce a novel regularization term coined \textit{NsLoss} that penalizes the classifier for high sensitivity of the network's neurons to input perturbations; 2) in order to push away the classifier's decision boundaries we use the well studied Jacobian regularization loss~\cite{hoffman2019robust, Jakubovitz_2018}.\\
Thus, we define a new training loss, $$L = CELoss + \lambda_1 \cdot NsLoss + \lambda_2 \cdot JacobRegLoss$$ where $CELoss$ is the standard Cross Entropy loss, $NsLoss$ is our newly introduced loss term and $JacobRegLoss$ is the Jacobian regularization loss.

We apply the proposed method on an already trained model by continuing to train it with the custom loss for a predefined number of epochs using a standard stochastic gradient descent based optimizer and a relatively low learning rate to allow the model to gain robustness without ``unlearning'' the classification task.

\subsection{\label{subsec:nsloss_computation}NsLoss Regularization Term}

We will now provide a detailed description of how to compute the NsLoss regularization term, as well as some intuition about its effectiveness in destroying the loss gradient landscape.

\subsubsection{NsLoss computation}
$NsLoss$ is computed by first computing the neural sensitivity using Algorithm~\ref{alg:compute_neural_sensitivity} to obtain: (1) $LayerSensitivity (LS)$  - a mapping of a model layer to a vector containing the sensitivity of each neuron in the layer, and (2) $MeanXLayerActivations (MLA)$ - a mapping of a model layer to a vector containing the average absolute value of the neuron activation over the current training batch. Then, to obtain the $NsLoss$, we compute a sum over all layers of the weighted sum of neuron sensitivity values weighted by the mean activation of the neuron to assign a larger weight to neurons with higher activation values. More formally, we compute: 
$$NsLoss \gets \sum_{L \in Layers}{LS[L] \bullet MLA[L]}$$

\vspace{-10pt}
\begin{algorithm}[h!]
\caption{Neural sensitivity computation}
\scriptsize
    \begin{algorithmic}[1]
        \State\textbf{Inputs:}
            \State $\qquad$ $M \gets$ model being trained
            \State $\qquad$ $X \gets$ batch of training samples
            \State $\qquad$ $NsEps \gets$ set $L_2$ radius in which to estimate sensitivity
            \State $\qquad$ $PG \gets$ Perturbation generator
            \State $\qquad$ $N \gets$ number of random perturbations to sample for each training sample
            \State $\qquad$ $Layers \gets$ list of model layers to compute sensitivity for
        \State\textbf{Outputs:}
            \State $\qquad$ $Layers Sensitivity \gets$ mapping of layer to vector of per neuron sensitivity scores
        
        \Procedure{ComputeNeuralSensitivity($M$, $X$, $NsEps$, $PG$, $N$, $Layers$)}{}
            \For{\texttt{$x \in X$}}
                \State{$X_p^{(i)}, i \in Z_N \gets$ Use $PG$ to generate $N$ random samples $x_1$, $x_2$, ..., $x_N$ with $\norm{x-x_j}_2 = NsEps$}
            \EndFor
            \State{$PerLayerXActivations \gets$ Evaluate $M(X)$ and store activations of each $layer \in Layers$}
            \For{$i \in Z_N$}
            \State{$PerLayerX_{p}Activations^{(i)}\gets$ Evaluate $M(X_p^{(i)})$ and store activations of each $layer \in Layers$}
            \EndFor
            \For{$layer L \in Layers$}
            \State{$PerLayerActivationDiffSums[L] \gets \sum_{i=1}^N{\abs{PerLayerX_{p}Activations^{(i)}[L] - PerLayerXActivations[L]}}$}
            \State{$MeanXLayerActivations[L] \gets \frac{\sum_{i=1}^{len(X)}{\abs{PerLayerXActivations[L]^{(i)}}}}{len(X)}$}\Comment{element-wise sum of activation vectors over the input samples in the batch}
            \State{$LayerSensitivity[L] \gets \frac{PerLayerActivationDiffSums[L]}{len(X) \cdot N \cdot NsEps \cdot len(L) \cdot MeanXLayerActivations[L]}$}
            \EndFor
        \State \Return $LayerSensitivity, MeanXLayerActivations$
        \EndProcedure
    \end{algorithmic}
\label{alg:compute_neural_sensitivity}
\end{algorithm}

\subsubsection{Intuition about the effectiveness of NsLoss regularization}
The NsLoss regularization term is constructed in a way that penalizes the model for big differences in the activations of both the output neurons (logits) and internal neurons of the model, originating from random input perturbations in a predefined $L_2$ ball around each training sample. This has the obvious effect of optimizing the model to minimize said differences. Thus, upon successful optimization, the model's logits are expected to become close to constant in the respective $L_2$ balls, causing flattening of the loss surface and bringing the loss gradients close to zero. Once aggregated over the entire training set during the training process this is expected to have the effect of zeroing the loss gradients in an $NsEps$ neighborhood of the entire data manifold, where $NsEps$ is the hyper-parameter specifying the radius of the $L_2$ ball from which random input perturbation are sampled during the computation of the NsLoss regularization term.

\subsection{Hyperparameters}
The following is a description of the hyperparameters used in our defense.
\begin{itemize}
    \item \textbf{N} - The number of perturbed samples to use to estimate the neuron sensitivity: We used a constant N=5 in all experiments.
    This parameter has a substantial impact on the performance as additional N forward propagations of the model are performed for each input sample, increasing the training time in a linear fashion.
    \item \textbf{NsEps} - The radius of the $L_2$ ball from which perturbations for neuron sensitivity computation are generated.
    \item \textbf{$\{R_i\}$} - The set of regularizers/penalty terms used to train the model out of:
    \begin{itemize}
        \item \textbf{full} - Regularize the sensitivity of the logits and all ReLU outputs in all layers of the model using the NsLoss term described in section \ref{subsec:nsloss_computation}
        \item \textbf{jacobian\_reg} - The Jacobian regularization loss as implemented in \cite{hoffman2019robust}.
    \end{itemize}
    \item \textbf{$\{\lambda_i\}$} - The weights of the loss terms. 
    For each regularization term $R_i$, a value of $\lambda_i$ that is too high would cause the respective term to outweigh the cross-entropy loss, resulting in a model with accuracy at the level of random guessing.
    A value of $\lambda_i$ that is too low, on the other hand, would have negligible effect on model robustness. 
    Thus, we strive to choose the largest value of $\lambda_i$ that doesn't hurt the model's cross-entropy loss and validation set accuracy.
    Specifically, we followed the protocol below to choose each $\lambda_i$
    \begin{enumerate}
        \item For a standard model, compute the values of $R_i$ on 10 random batches from the training and validation sets and store the average value as $R_{i_0}$.
        \item Choose $\lambda_{i_0} = \frac{\log_2{NumClasses}}{R_{i_0}}$.
        \item Perform a binary search by selecting values of $\lambda_i$ that are larger and smaller than $\lambda_{i_0}$ but in the same order of magnitude. For each such $\lambda_i$, start training the model for an epoch. If the training cross-entropy loss reaches the cross-entropy of a random guess ($\log_2{NumClasses}$), then $\lambda_i$ is too high; otherwise it can be increased further. 
    \end{enumerate}
\end{itemize}

\section{\label{sec:evaluation}Evaluation}
\subsection{\label{subsec:robustness_eval_method}Robustness Evaluation Methodology}
\textbf{Threat model.} We evaluate the robustness of models trained using our defensive method in a white-box threat model, i.e., one in which the attacker has full access to the trained model, as well as information about the methods and parameters used to train the model. In this work we limit ourselves to $L_2$ attacks for the sake of simplicity, but the method can be naturally extended to defend against attacks targeting $L_0$, $L_\infty$ and other norms as well by aligning the sampling of random perturbations used in the computation of the NsLoss regularization term with the norm the adversarial attack is operating on, and possibly combining regularization terms targeting multiple norms.

\noindent\textbf{White-box evaluation best practices.} 
It is well known in the field that defense evaluation requires testing the model against a wide range of attacks, the most prominent of which are: (1) generic gradient-based (white-box) attacks; (2)  boundary-based gradient free (black-box) attacks; (3) transferability attacks, i.e., evaluating the robustness to adversarial examples created against an undefended, substitute model; and (4) adaptive attacks, i.e., attacks that are tailored to defeat the defenses using a suitable choice of attack configuration parameters, as well as customizing state-of-the-art attacks to make them more effective against the specific defense~\cite{carlini2019evaluating,tramer2020adaptive,carlini2017adversarial,athalye2018obfuscated}.
Specifically, we evaluate our defense's capabilities in the face of the following attacks:
\begin{enumerate}
\item\textit{Generic gradient-based attacks.}
    We use the PGD and C\&W attacks with the $L_2$ norm. 
    For PGD, we use a strong attack configuration with a large number of iterations. 
    For C\&W, we use %
    different number of iterations and target confidence values (see Table~\ref{tab:all-attacks} for details).
    \item\textit{Boundary-based attacks.} We use the strong HopSkipJump attack method with the $L_2$ norm.
    \item\textit{Transferability attacks.} We evaluate the transferability of all of the above attacks by creating adversarial examples against %
    an undefended model, and evaluating the robust model on them to compute the adversarial test accuracy.
    \item\textit{Adaptive attacks.} In contrast to most defense methods that succumbed to approaches, such as backward pass differentiable approximation (BPDA)~\cite{athalye2018obfuscated}, expectation over transformation (EOT)~\cite{athalye2018obfuscated}, and variations of the C\&W attack with a customized loss function~\cite{carlini2017adversarial}, our approach is inherently difficult to attack adaptively as it lacks any explicit inference time mechanisms that the attacker needs to overcome. 
    To challenge our defense, we use two strengthened variations of the C\&W attack:
    \begin{itemize}
        \item \textit{Attack restarts with random initialization} - The standard C\&W implementation searches for a minimal perturbation starting with an all-zero perturbation vector. Our variation starts with a uniformly sampled perturbation of an $L_2$ norm given by an input parameter. We perform a configurable number of restarts of the attack, each with a different random initial perturbation. This attack aims at overcoming a flat gradient surface in the vicinity of an attacked input sample by ``jumping'' out of the gradient valley/plateau and then starting the gradient-based search.
        \item \textit{Gradient expectation over randomness} - We modified the standard C\&W gradient computation with one that averages the gradient values over a configurable sized set of inputs chosen at random from an $L_2$ ball with configurable radius around the current input sample. 
        This attack aims at overcoming a noisy gradient surface by smoothing the gradient values used for the attack.
    \end{itemize}
\end{enumerate}

\noindent\textbf{Evaluation metrics}
We measure the robustness of the evaluated models using the well established \textit{adversarial test accuracy} metric. This metric is calculated by executing an adversarial attack on the entire test set (or a subset of it) and computing the accuracy of the model on the set of samples generated by the attack, which are a mix of adversarial examples (where an attack succeeds) and normal samples (where the attack fails).
In order to provide a deeper understanding of the model's robustness, the adversarial test accuracy is measured for a wide range of perturbation budgets (i.e., the L2 norm between the original sample and the resulting adversarial sample).

\subsection{Evaluation Setup}
\textbf{Datasets and base models.}
Conforming to common practice in the field, we evaluate the defense on the classic MNIST \cite{lecun1998gradient} and CIFAR-10~\cite{krizhevsky2009learning} image classification datasets.
For CIFAR-10, we use a pretrained VGG19 model~\cite{simonyan2014deep} which achieves clean test accuracy of 92.43\%, whereas for MNIST, we use a model architecture taken from the Keras MNIST example (https://keras.io/examples/mnist\_cnn/) which achieves clean test accuracy of 99.25\%.

\noindent\textbf{Optimization and hyperparameters.}
All of the robust models were created by starting with a pretrained model and retraining with the configured set of regularization terms.
Specifically, we trained the models using the Adam optimizer at a low learning rate of $1e-4$. For MNIST, we trained for 20 epochs, and for CIFAR-10, we trained for 100 epochs.
Table~\ref{tab:defense_models_hyperparam} summarizes the configurations of the robust models' hyperparameters.

\noindent\textbf{Hardware.}
All experiments were executed on a dedicated server with 8 x NVIDIA RTX 2080 Ti (11GB RAM) GPUs. 

\begin{table}[t]
\centering
\scriptsize
\begin{tabular}{|l|l|cccc|c|}
\hline
\multirow{2}{*}{Dataset} & \multirow{2}{*}{Defense} & \multicolumn{4}{c|}{Hyperparameters} & \multicolumn{1}{c|}{Add. Info}                                                  \\
                         &            & $\lambda_i$   & $\epsilon$   & N    &  epoch  & ACC \\
\hline
\multirow{3}{*}{MNIST}    & NsLoss    & 2.0           & 1.0          & 5    &   20  & 95.65 \\
                          & JacobReg  & 0.01          & N/A          & N/A  &   20  & 94.95 \\
                          & Combined  & 2.0,0.01      & 1.0          & 5    &   20  & 96.22 \\
\hline
\multirow{3}{*}{CIFAR10}  & NsLoss    & 10.0          & 3.0          & 10   &   100  & 88.96 \\
                          & JacobReg  & 0.1           & N/A          & N/A  &   100  & 87.60 \\
                          & Combined  & 10.0,0.01      & 3.0          & 10   &   100  & 89.02 \\
\hline
\end{tabular}
\caption{Model hyperparameters}
\label{tab:defense_models_hyperparam}
\end{table}

\begin{table*}[t!]
\scriptsize
\centering
\begin{tabular}{|l|l|c|c|c|l|}
\hline
Type                     & Attack Name          & D         & Add. Info      & Impl. & Parameters \\
\hline
\multirow{5}{*}{Gradient} & \multirow{3}{*}{C\&W}  & $L_2$     & Standard             & foolbox        &  \begin{tabular}[c]{@{}l@{}}steps=1,000; stepsize=\{CIFAR10:0.01,MNIST: 0.1\};\\ initial\_const=\{CIFAR10:1e-3,MNIST: 10\}; binary\_search\_steps=9; \textbf{confidence=0}\end{tabular} \\
                                &                      & $L_2$     & High Conf      & foolbox        & \begin{tabular}[c]{@{}l@{}}steps=1,000; stepsize=\{CIFAR10:0.01,MNIST: 0.1\};\\ initial\_const=\{CIFAR10:1e-3, MNIST: 10\}; binary\_search\_steps=9; \textbf{confidence=5}\end{tabular} \\
                                &                      & $L_2$     & Strong               & foolbox        & \begin{tabular}[c]{@{}l@{}}\textbf{steps=10,000}; stepsize=\{CIFAR10:0.01, MNIST: 0.1\};\\ initial\_const=\{CIFAR10:1e-3, MNIST: 10\}; binary\_search\_steps=9; \textbf{confidence=0}\end{tabular} \\
                                & \multirow{1}{*}{PGD} & $L_2$     & Random Init          & advertorch     & rand\_init=True; eps\_ter=0.01; nb\_ter=1,000    \\
\hline
Boundary                  & HopSkipJump & $L_2$     &                      & ART        & init\_val=100; \textbf{init\_size=1,000}; max\_ter=50; \\
\hline
\multirow{2}{*}{Adaptive}       & C\&W with Restarts        & $L_2$     & Random init & foolbox        & number of restarts = 400 \\
                                & C\&W with Grad-Exp                  & $L_2$     &     Based on C\&W        & custom         & steps=1,000; confidence=0; ge\_sample\_count=10; ge\_eps=0.01 \\
\hline
\end{tabular}
\caption{Detailed attack parameters}
\label{tab:all-attacks}
\end{table*}

\noindent\textbf{Attack configurations.}
For \textit{untargeted} attacks, we randomly selected 1,024 samples from the respective \textit{test} set to serve as the attack inputs, whereas for \textit{targeted} attacks, we randomly selected 113 samples from the respective \textit{test} set, and launched an attack targeting each possible target class (except for the real class), for a total of 1,017 attack instances (both MNIST and CIFAR-10 have 10 target classes, thus we have 10-1=9 attack instances per sample).
To be sure that the results of the attacks are comparable across different models and attack types, we set a constant random seed with the meaningful value of \textit{42} for each attack sequence (to ensure that the same random samples are chosen each time).

Table~\ref{tab:all-attacks} presents the attack types and parameters. For each attack we use both the targeted and untargeted variants, where applicable. We used state-of-the-art, open-source implementations of the various attacks using the following popular frameworks: \textit{Foolbox}~\cite{rauber2017foolbox}, \textit{Advertorch}~\cite{ding2019advertorch}, and \textit{Adversarial Robustness Toolbox} (ART)~\cite{art2018}. For all the attacks we used $L_2$ perturbation budgets of 0.01, 0.1, 0.2, 0.5, 1.0, 2.0, 3.0, 5.0, 10.0, and 20.0.

\subsection{Results}
\begin{figure*}[t]
\centering
\begin{subfigure}{1.0\textwidth}
  \centering
  \includegraphics[width=1.0\linewidth]{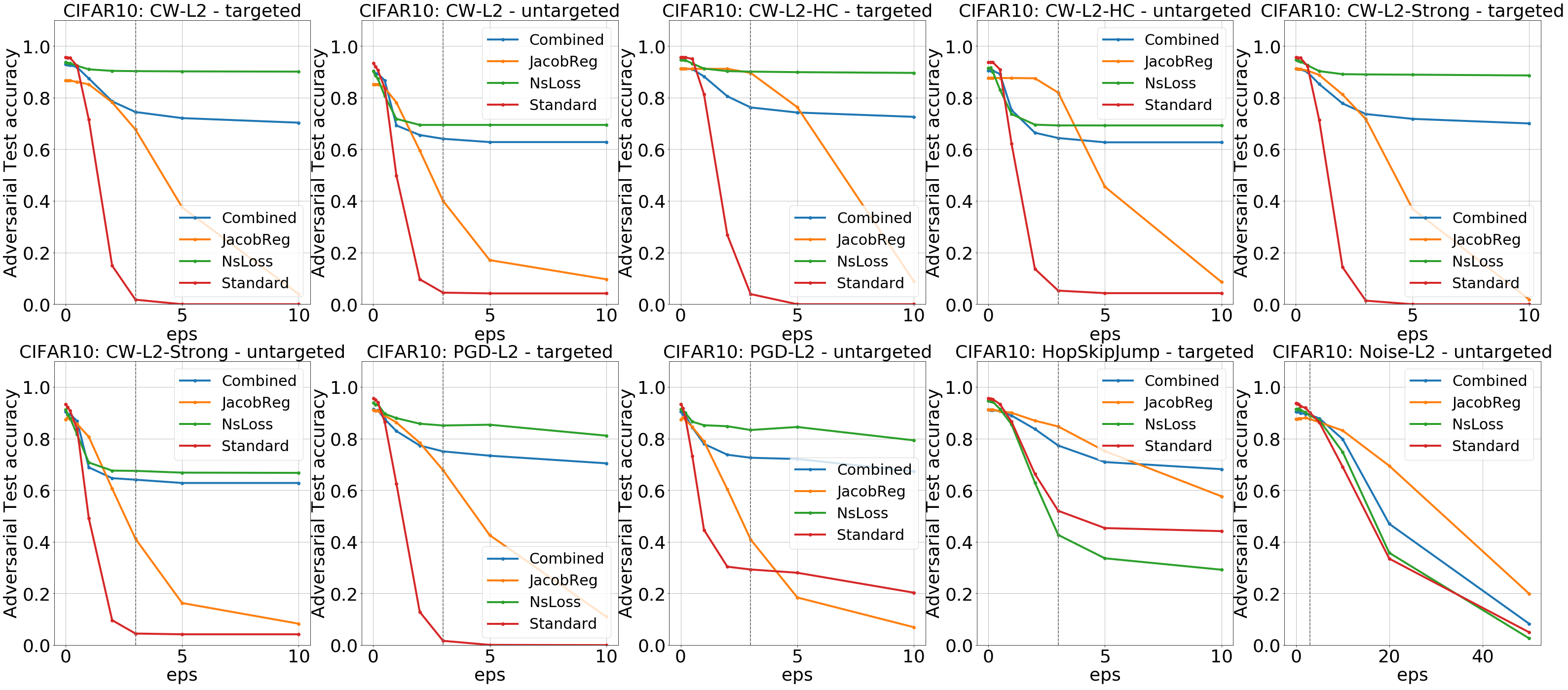}  
  \caption{CIFAR10 dataset}
  \label{fig:cifar10-attack-results}
\end{subfigure}

\begin{subfigure}{1.0\textwidth}
  \centering
  \includegraphics[width=1.0\linewidth]{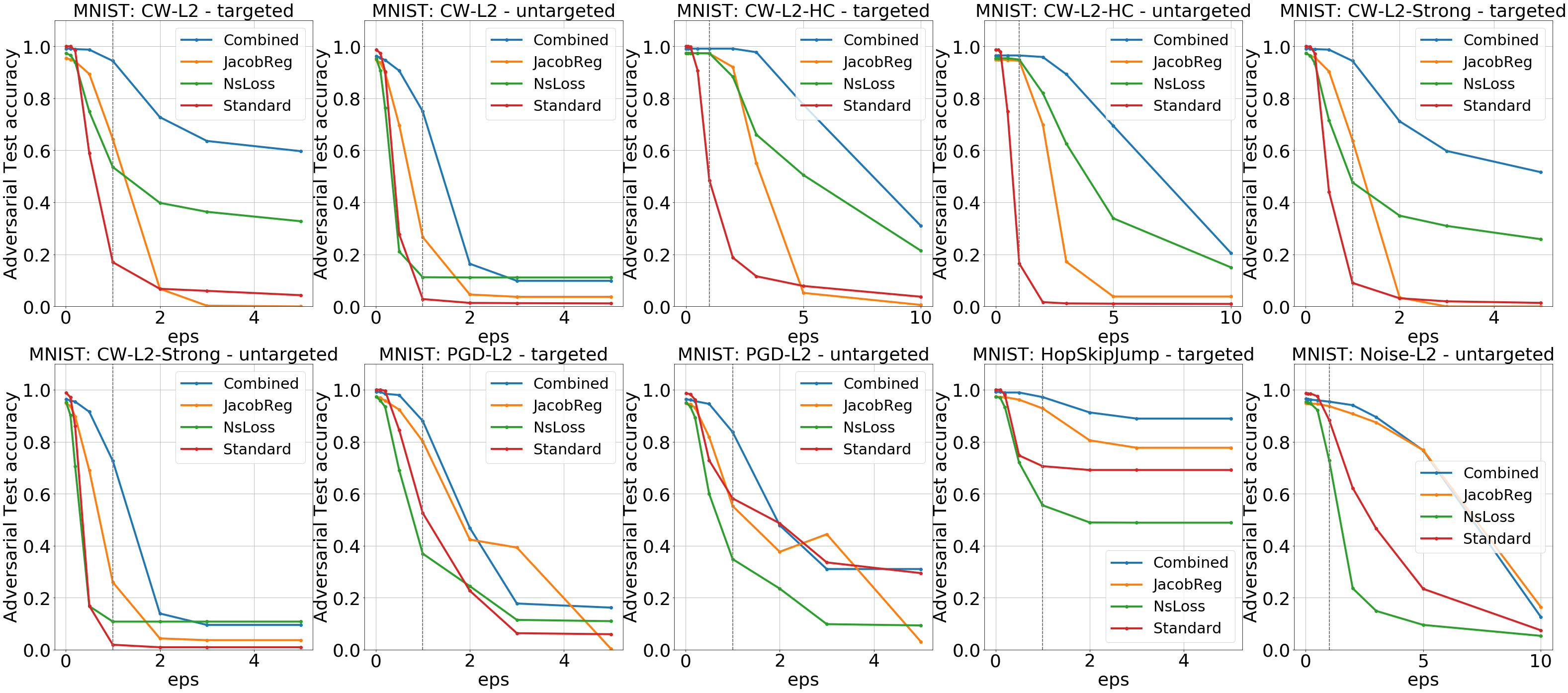}  
  \caption{MNIST dataset}
  \label{fig:mnist-attack-results}
\end{subfigure}
\caption{Gradient-based and boundary-based attack results}
\label{fig:attack_results}
\end{figure*}

\textbf{Gradient-based and boundary-based attacks.}
Tables~\ref{tab:attack_res_mnist} and~\ref{tab:attack_res_cifar10} contain a summary of the results of all generic gradient and boundary-based attacks, represented by the adversarial test accuracy, for two different perturbation budgets ($\epsilon$): a small budget that is at the boundary of perturbations perceivable by the naked human eye and a large one to showcase the model's ability to cope with very large perturbations. 
In addition, in Figures~\ref{fig:cifar10-attack-results} and \ref{fig:mnist-attack-results} we present complete plots of adversarial test accuracy vs. perturbation budget ($\epsilon$) for a few representative attacks.

These results clearly support our three main hypotheses, namely: (1) The \textit{NsLoss} regularization is highly effective in thwarting gradient-based attacks, even for very large perturbation budgets, as can be seen by the high test accuracy of both the the \textit{NsLoss} and \textit{Combined} models for the gradient-based attacks (PGD and CW variations). 
(2) The Jacobian regularization is highly effective in thwarting boundary attacks, as can be seen by the high test accuracy of both the \textit{JacobReg} and \textit{Combined} models.
(3) Using a combined loss preserves the benefits of both the \textit{NsLoss} and \textit{JacobReg}, albeit with slightly lower accuracy than the use of \textit{NsLoss} alone, as can be seen by comparing \textit{Combined} against \textit{NsLoss} for the gradient attacks and against Jacobian regularization for the HopSkipJump boundary attack.

\begin{table}[h]
\centering
\scriptsize
\begin{tabular}{|l|l|cc|cc|cc|cc|}
\hline
\multirow{2}{*}{}           & \multirow{2}{*}{Attack} & \multicolumn{2}{c|}{Standard} & \multicolumn{2}{c|}{JacobReg} & \multicolumn{2}{c|}{NsLoss} & \multicolumn{2}{c|}{Combined} \\
                            &                         & $\epsilon_1$           & $\epsilon_{5}$         & $\epsilon_1$           & $\epsilon_{5}$         & $\epsilon_1$          & $\epsilon_{5}$        & $\epsilon_1$           & $\epsilon_{5}$         \\
\hline
\multirow{4}{*}{\rotatebox{90}{Untargeted}} & CW-L2                   & 3               & 1         & 27          & 4         & 11         & 11        & 75          & 10         \\
                            & CW-L2-HC                & 17          & 1         & 94          & 4         & 95         & 34        & 96          & 69         \\
                            & CW-L2-Strong            & 2          & 1         & 26          & 4         & 11         & 11        & 73          & 10         \\
                            & PGD-L2                  & 58          & 29         & 55          & 3         & 35         & 9        & 84          & 31         \\
\hline
\multirow{5}{*}{\rotatebox{90}{Targeted}}   & CW-L2                   & 17          & 4         & 54          & 33         & 64         & 0        & 94          & 60         \\
                            & CW-L2-HC                & 48          & 8         & 97          & 5         & 97         & 51        & 99          & 78         \\
                            & CW-L2-Strong            & 9          & 1         & 64          & 0         & 48         & 26        & 94          & 52         \\
                            & PGD-L2                  & 53          & 6         & 80          & 0         & 37         & 11        & 88          & 16         \\
\cline{2-10}
                            & HopSkipJump             & 71          & 69         & 93          & 78         & 56         & 49        & 97          & 89        \\
\hline
\end{tabular}
\caption{MNIST - adversarial test accuracy}
\label{tab:attack_res_mnist}
\end{table}

\begin{table}[h]
\centering
\scriptsize
\begin{tabular}{|l|l|cc|cc|cc|cc|}
\hline
\multirow{2}{*}{}           & \multirow{2}{*}{Attack} & \multicolumn{2}{c|}{Standard} & \multicolumn{2}{c|}{JacobReg} & \multicolumn{2}{c|}{NsLoss} & \multicolumn{2}{c|}{Combined} \\
                            &                         & $\epsilon_3$           & $\epsilon_{10}$         & $\epsilon_3$           & $\epsilon_{10}$         & $\epsilon_3$          & $\epsilon_{10}$        & $\epsilon_3$           & $\epsilon_{10}$         \\
\hline
\multirow{4}{*}{\rotatebox{90}{Untargeted}} & CW-L2                   & 4          & 4         & 40          & 10         & 69         & 69        & 64          & 63         \\
                            & CW-L2-HC                & 5          & 4         & 82          & 9         & 69         & 69        & 64          & 63         \\
                            & CW-L2-Strong            & 4          & 4         & 41          & 8         & 67         & 67        & 64          & 63         \\
                            & PGD-L2                  & 29          & 20         & 41          & 7         & 83         & 79        & 73          & 67         \\
\hline
\multirow{5}{*}{\rotatebox{90}{Targeted}}   & CW-L2                   & 2          & 0         & 68          & 4         & 90         & 90        & 74          & 70         \\
                            & CW-L2-HC                & 4          & 0         & 90          & 9         & 90         & 90        & 76          & 73         \\
                            & CW-L2-Strong            & 1          & 0         & 72          & 2         & 89         & 89        & 74          & 70         \\
                            & PGD-L2                  & 2          & 0         & 68          & 11         & 85         & 81        & 75          & 70         \\
\cline{2-10}
                            & HopSkipJump             & 52          & 44         & 85          & 58         & 43         & 29        & 77          & 68        \\
\hline
\end{tabular}
\caption{CIFAR10 - adversarial test accuracy}
\label{tab:attack_res_cifar10}
\end{table}

\begin{figure*}[t]
\centering
\begin{subfigure}{1.0\textwidth}
  \centering
  \includegraphics[width=1.0\linewidth]{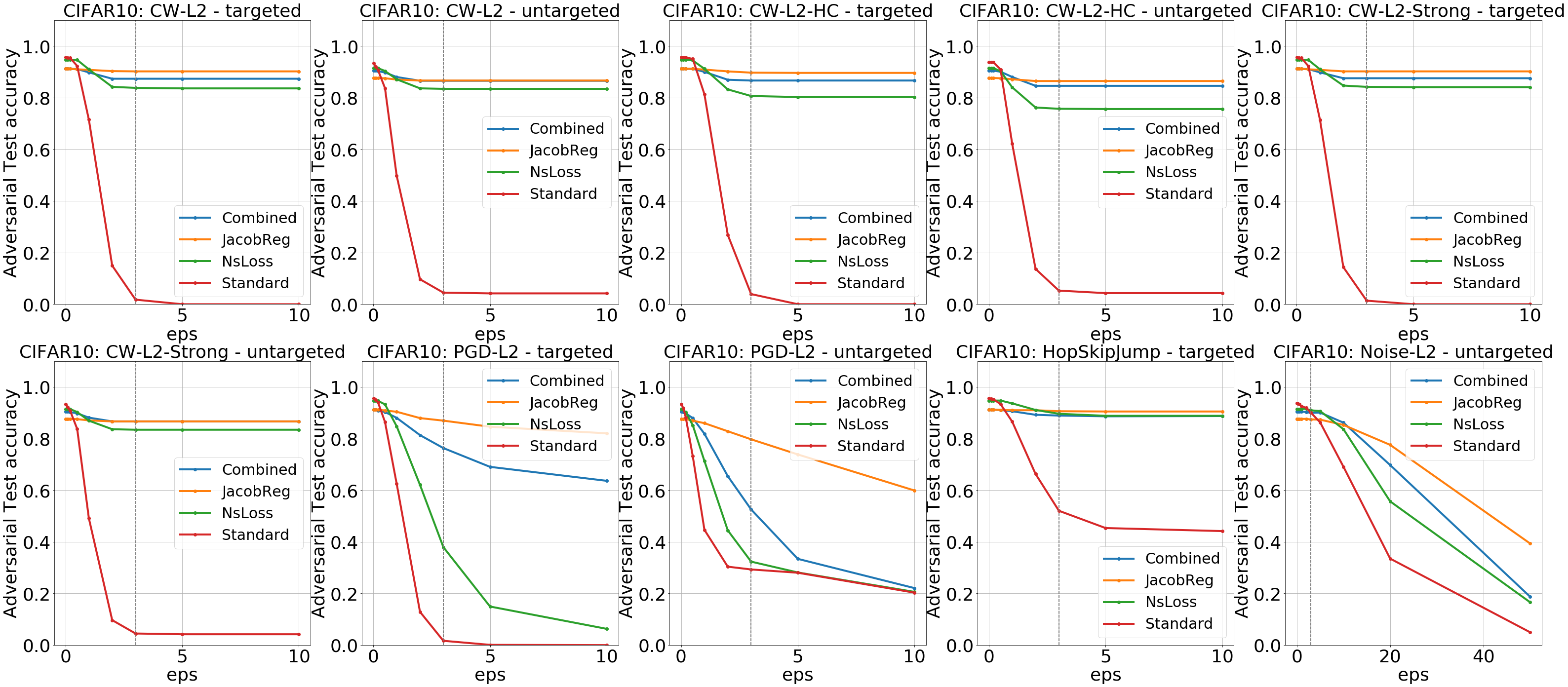}  
  \caption{CIFAR10 dataset}
  \label{fig:cifar10-transfer-attack-results}
\end{subfigure}

\begin{subfigure}{1.0\textwidth}
  \centering
  \includegraphics[width=1.0\linewidth]{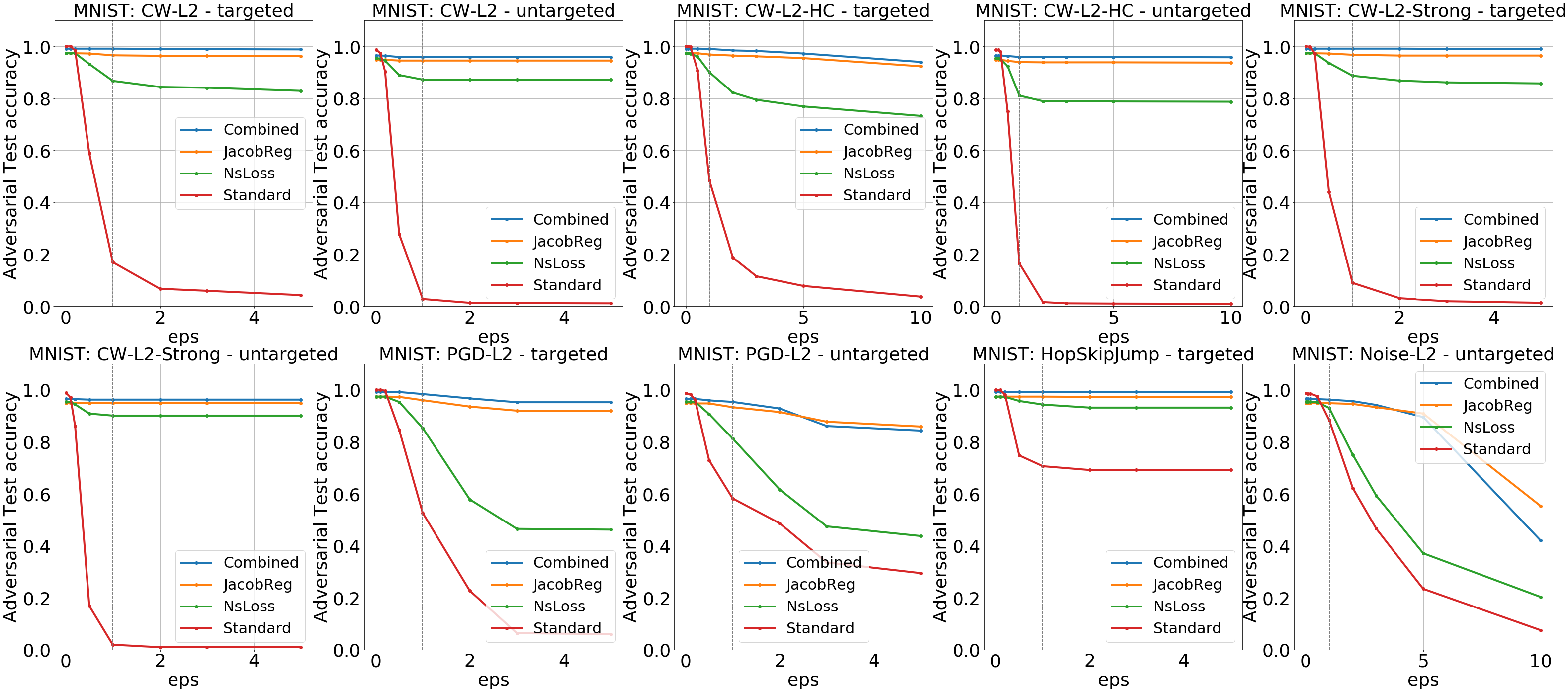}  
  \caption{MNIST dataset}
  \label{fig:mnist-transfer-attack-results}
\end{subfigure}
\caption{Transferability attack results}
\label{fig:transfer_attack_results}
\end{figure*}

\noindent\textbf{Transferability attacks.}
Tables~\ref{tab:attack_res_transfer_mnist} and~\ref{tab:attack_res_transfer_cifar10} and figure~\ref{fig:transfer_attack_results} summarize the adversarial test accuracy against adversarial examples created using the standard (undefended) model. 
The high test accuracy scores show that a transferability attack doesn't work against the Jacobian regularization-based models, the NsLoss models, or combined models.

\begin{table}[h]
\centering
\scriptsize
\begin{tabular}{|l|l|cc|cc|cc|cc|}
\hline
\multirow{2}{*}{}           & \multirow{2}{*}{Attack} & \multicolumn{2}{c|}{Standard} & \multicolumn{2}{c|}{JacobReg} & \multicolumn{2}{c|}{NsLoss} & \multicolumn{2}{c|}{Combined} \\
                            &                         & $\epsilon_1$           & $\epsilon_{5}$         & $\epsilon_1$           & $\epsilon_{5}$         & $\epsilon_1$          & $\epsilon_{5}$        & $\epsilon_1$           & $\epsilon_{5}$         \\
\hline
\multirow{4}{*}{\rotatebox{90}{Untargeted}} & CW-L2                   & 3          & 1         & 95          & 95         & 87         & 87        & 96          & 96         \\
                            & CW-L2-HC                & 17          & 1         & 94          & 94         & 81         & 79        & 96          & 96         \\
                            & CW-L2-Strong            & 2          & 1         & 95          & 95         & 90         & 90        & 96          & 96         \\
                            & PGD-L2                  & 58          & 29         & 93          & 86         & 81         & 44        & 95          & 84         \\
\hline
\multirow{5}{*}{\rotatebox{90}{Targeted}}   & CW-L2                   & 17          & 4         & 97          & 96         & 87         & 83        & 99          & 99         \\
                            & CW-L2-HC                & 48          & 8         & 97          & 95         & 90         & 77        & 99          & 97         \\
                            & CW-L2-Strong            & 9          & 1         & 97          & 96         & 89         & 86        & 99          & 99         \\
                            & PGD-L2                  & 53          & 6         & 96          & 92         & 85         & 46        & 98          & 95         \\
\cline{2-10}
                            & HopSkipJump             & 71          & 69         & 97          & 97         & 94         & 93        & 99          & 99        \\
\hline
\end{tabular}
\caption{MNIST - adversarial test accuracy with transferability attack}
\label{tab:attack_res_transfer_mnist}
\end{table}

\begin{table}[h]
\centering
\scriptsize
\begin{tabular}{|l|l|cc|cc|cc|cc|}
\hline
\multirow{2}{*}{}           & \multirow{2}{*}{Attack} & \multicolumn{2}{c|}{Standard} & \multicolumn{2}{c|}{JacobReg} & \multicolumn{2}{c|}{NsLoss} & \multicolumn{2}{c|}{Combined} \\
                            &                         & $\epsilon_3$           & $\epsilon_{10}$         & $\epsilon_3$           & $\epsilon_{10}$         & $\epsilon_3$          & $\epsilon_{10}$        & $\epsilon_3$           & $\epsilon_{10}$         \\
\hline
\multirow{4}{*}{\rotatebox{90}{Untargeted}} & CW-L2                   & 4          & 4         & 87          & 87         & 83         & 83        & 87          & 87         \\
                            & CW-L2-HC                & 5          & 4         & 86          & 86         & 76         & 76        & 85          & 85         \\
                            & CW-L2-Strong            & 4          & 4         & 87          & 87         & 83         & 83        & 87          & 87         \\
                            & PGD-L2                  & 29          & 20         & 80          & 60         & 32         & 21        & 53          & 22          \\
\hline
\multirow{5}{*}{\rotatebox{90}{Targeted}}   & CW-L2                   & 2          & 0         & 90          & 90         & 84         & 84        & 87          & 87         \\
                            & CW-L2-HC                & 4          & 0         & 90          & 90         & 81         & 80       & 87          & 87         \\
                            & CW-L2-Strong            & 1          & 0         & 90          & 90         & 84         & 84        & 88          & 88         \\
                            & PGD-L2                  & 2          & 0         & 87          & 82         & 38         & 6        & 76          & 64          \\
\cline{2-10}
                            & HopSkipJump             & 52          & 44         & 91          & 90          & 90         & 89        & 89          & 89        \\
\hline
\end{tabular}
\caption{CIFAR10 - adversarial test accuracy with transferability attack}
\label{tab:attack_res_transfer_cifar10}
\end{table}

\begin{figure}[h]
\centering
\begin{subfigure}{0.45\textwidth}
  \centering
  \includegraphics[width=1.0\linewidth]{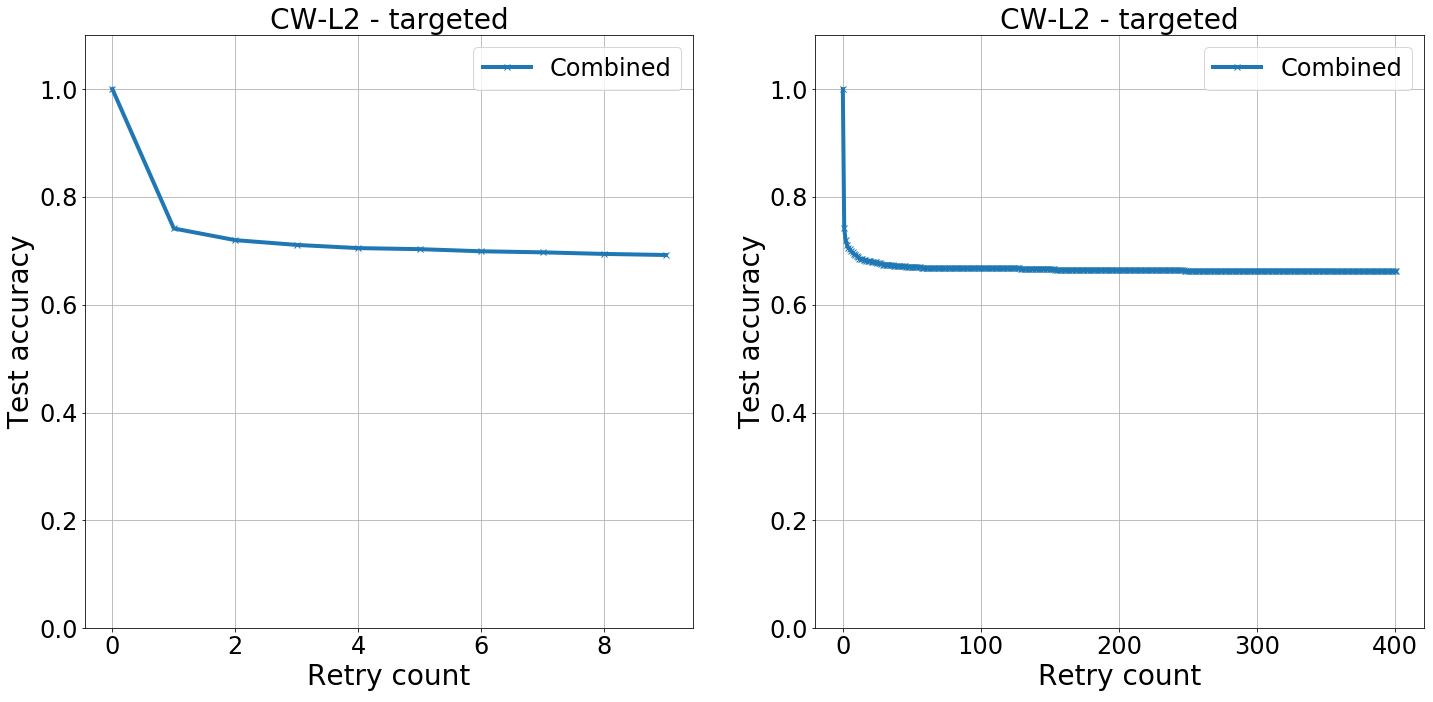}  
  \caption{C\&W with random restart attack}
  \hspace{20pt}
  \label{fig:cifar10-random-restart}
\end{subfigure}

\begin{subfigure}{0.45\textwidth}
  \centering
  \includegraphics[width=1.0\linewidth]{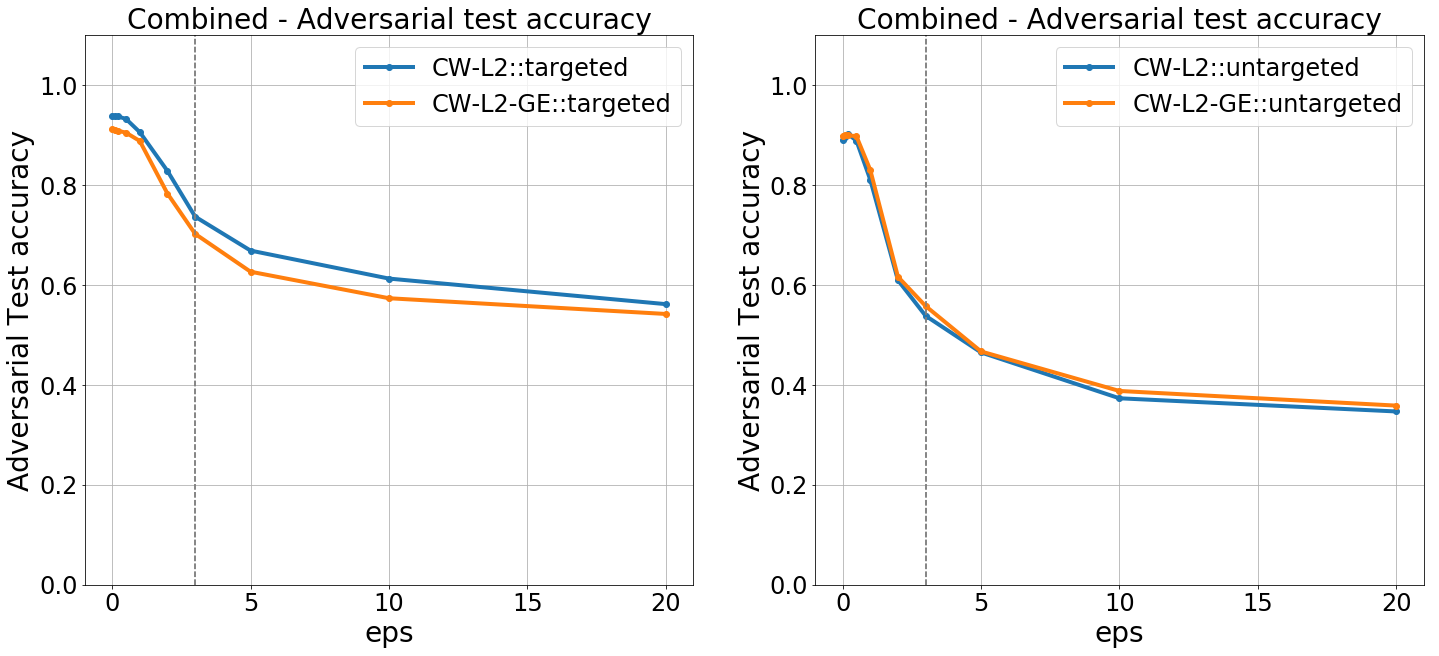}  
  \caption{C\&W with gradient expectation over randomness}
  \label{fig:cifar10-ge-attack}
\end{subfigure}
\caption{Adaptive attack results on CIFAR10 dataset}
\label{fig:adaptive_attack_results}
\end{figure}

\noindent\textbf{Adaptive attacks.}
Figure~\ref{fig:cifar10-random-restart} presents the adversarial test accuracy of our \textit{combined} model as a function of the number of CW-L2 targeted attack restarts. 
We obtained very similar results for the untargeted CW-L2 attack. 
This can be clearly seen, except for the slight decrease in test accuracy observed during the first few attack restarts; note that up to 400 additional restarts were shown to have a negligible affect on the test accuracy which implies that the model is robust to this kind of attack.

Figure~\ref{fig:cifar10-ge-attack} shows the results of the C\&W L2 with \textit{gradient expectation over randomness} attacks and a standard C\&W L2 attack. 
It is clearly seen that the gradient expectation over randomness has a negligible effect on the attack success rate, which means that the model is robust to this adaptive attack as well.

\section{\label{sec:conclusion}Conclusions and future work}
The results of our experiments validate the effectiveness of the proposed method in making the evaluated models robust against a white-box adversary with full access to the model and full knowledge of the defensive method. The proposed defense is unique in that it successfully combines both state of the art defensive approaches, namely: 1) Flattening the loss gradients near the training samples to hinder white-box attacks and 2) Pushing decision boundaries away from the training data manifolds to weaken the adversarial nature of adversarial examples found using boundary based attacks by effectively increasing their perturbation distance from valid inputs. Although recent work in the field suggests that the existence of adversarial examples is inevitable, this work presents an effective practical approach to greatly increase the difficulty of finding them on the one hand, and weaken their adversarial nature on the other hand.

The present work evaluates the proposed defensive method on popular image classification benchmark datasets. However, the method is generic and a possible extension of this research is to apply it to additional domains and neural network architectures.

\newpage
\clearpage
\fontsize{9.0pt}{10.0pt} \selectfont

\bibliographystyle{IEEEtran}
\bibliography{references}

\begin{thebibliography}{10}
\providecommand{\url}[1]{#1}
\csname url@samestyle\endcsname
\providecommand{\newblock}{\relax}
\providecommand{\bibinfo}[2]{#2}
\providecommand{\BIBentrySTDinterwordspacing}{\spaceskip=0pt\relax}
\providecommand{\BIBentryALTinterwordstretchfactor}{4}
\providecommand{\BIBentryALTinterwordspacing}{\spaceskip=\fontdimen2\font plus
\BIBentryALTinterwordstretchfactor\fontdimen3\font minus
  \fontdimen4\font\relax}
\providecommand{\BIBforeignlanguage}[2]{{%
\expandafter\ifx\csname l@#1\endcsname\relax
\typeout{** WARNING: IEEEtran.bst: No hyphenation pattern has been}%
\typeout{** loaded for the language `#1'. Using the pattern for}%
\typeout{** the default language instead.}%
\else
\language=\csname l@#1\endcsname
\fi
#2}}
\providecommand{\BIBdecl}{\relax}
\BIBdecl

\bibitem{shafahi2018adversarial}
A.~Shafahi, W.~R. Huang, C.~Studer, S.~Feizi, and T.~Goldstein, ``Are
  adversarial examples inevitable?'' \emph{arXiv preprint arXiv:1809.02104},
  2018.

\bibitem{cohen2019certified}
J.~M. Cohen, E.~Rosenfeld, and J.~Z. Kolter, ``Certified adversarial robustness
  via randomized smoothing,'' \emph{arXiv preprint arXiv:1902.02918}, 2019.

\bibitem{shamir2019simple}
A.~Shamir, I.~Safran, E.~Ronen, and O.~Dunkelman, ``A simple explanation for
  the existence of adversarial examples with small hamming distance,''
  \emph{arXiv preprint arXiv:1901.10861}, 2019.

\bibitem{katzir2020gradients}
Z.~Katzir and Y.~Elovici, ``Gradients cannot be tamed: Behind the impossible
  paradox of blocking targeted adversarial attacks,'' \emph{IEEE Transactions
  on Neural Networks and Learning Systems}, 2020.

\bibitem{goodfellow2014explaining}
I.~J. Goodfellow, J.~Shlens, and C.~Szegedy, ``Explaining and harnessing
  adversarial examples,'' \emph{arXiv preprint arXiv:1412.6572}, 2014.

\bibitem{kurakin2016adversarial}
A.~Kurakin, I.~Goodfellow, and S.~Bengio, ``Adversarial machine learning at
  scale,'' 2016.

\bibitem{goodfellow2016dlbook}
I.~Goodfellow, Y.~Bengio, and A.~Courville, \emph{Deep Learning}.\hskip 1em
  plus 0.5em minus 0.4em\relax MIT Press, 2016,
  \url{http://www.deeplearningbook.org}.

\bibitem{carlini2017towards}
N.~Carlini and D.~Wagner, ``Towards evaluating the robustness of neural
  networks,'' in \emph{2017 IEEE Symposium on Security and Privacy (SP)}.\hskip
  1em plus 0.5em minus 0.4em\relax IEEE, 2017, pp. 39--57.

\bibitem{papernot2017practical}
N.~Papernot, P.~McDaniel, I.~Goodfellow, S.~Jha, Z.~B. Celik, and A.~Swami,
  ``Practical black-box attacks against machine learning,'' in
  \emph{Proceedings of the 2017 ACM on Asia conference on computer and
  communications security}.\hskip 1em plus 0.5em minus 0.4em\relax ACM, 2017,
  pp. 506--519.

\bibitem{chen2019hopskipjumpattack}
J.~Chen, M.~I. Jordan, and M.~J. Wainwright, ``Hopskipjumpattack: A
  query-efficient decision-based attack,'' 2019.

\bibitem{Xiao2020Enhancing}
\BIBentryALTinterwordspacing
C.~Xiao, P.~Zhong, and C.~Zheng, ``Enhancing adversarial defense by
  k-winners-take-all,'' in \emph{International Conference on Learning
  Representations}, 2020. [Online]. Available:
  \url{https://openreview.net/forum?id=Skgvy64tvr}
\BIBentrySTDinterwordspacing

\bibitem{pang2019rethinking}
T.~Pang, K.~Xu, Y.~Dong, C.~Du, N.~Chen, and J.~Zhu, ``Rethinking softmax
  cross-entropy loss for adversarial robustness,'' 2019.

\bibitem{Sen2020EMPIR}
\BIBentryALTinterwordspacing
S.~Sen, B.~Ravindran, and A.~Raghunathan, ``Empir: Ensembles of mixed precision
  deep networks for increased robustness against adversarial attacks,'' in
  \emph{International Conference on Learning Representations}, 2020. [Online].
  Available: \url{https://openreview.net/forum?id=HJem3yHKwH}
\BIBentrySTDinterwordspacing

\bibitem{hoffman2019robust}
J.~Hoffman, D.~A. Roberts, and S.~Yaida, ``Robust learning with jacobian
  regularization,'' 2019.

\bibitem{Jakubovitz_2018}
\BIBentryALTinterwordspacing
D.~Jakubovitz and R.~Giryes, ``Improving dnn robustness to adversarial attacks
  using jacobian regularization,'' \emph{Lecture Notes in Computer Science}, p.
  525–541, 2018. [Online]. Available:
  \url{http://dx.doi.org/10.1007/978-3-030-01258-8_32}
\BIBentrySTDinterwordspacing

\bibitem{athalye2018obfuscated}
A.~Athalye, N.~Carlini, and D.~Wagner, ``Obfuscated gradients give a false
  sense of security: Circumventing defenses to adversarial examples,'' 2018.

\bibitem{carlini2017adversarial}
N.~Carlini and D.~Wagner, ``Adversarial examples are not easily detected:
  Bypassing ten detection methods,'' in \emph{Proceedings of the 10th ACM
  Workshop on Artificial Intelligence and Security}.\hskip 1em plus 0.5em minus
  0.4em\relax ACM, 2017, pp. 3--14.

\bibitem{tramer2020adaptive}
F.~Tramer, N.~Carlini, W.~Brendel, and A.~Madry, ``On adaptive attacks to
  adversarial example defenses,'' \emph{arXiv preprint arXiv:2002.08347}, 2020.

\bibitem{drucker1992improving}
H.~Drucker and Y.~Le~Cun, ``Improving generalization performance using double
  backpropagation,'' \emph{IEEE Transactions on Neural Networks}, vol.~3,
  no.~6, pp. 991--997, 1992.

\bibitem{lecun1998gradient}
Y.~LeCun, L.~Bottou, Y.~Bengio, P.~Haffner \emph{et~al.}, ``Gradient-based
  learning applied to document recognition,'' \emph{Proceedings of the IEEE},
  vol.~86, no.~11, pp. 2278--2324, 1998.

\bibitem{krizhevsky2009learning}
A.~Krizhevsky, G.~Hinton \emph{et~al.}, ``Learning multiple layers of features
  from tiny images,'' Citeseer, Tech. Rep., 2009.

\bibitem{carlini2019evaluating}
N.~Carlini, A.~Athalye, N.~Papernot, W.~Brendel, J.~Rauber, D.~Tsipras,
  I.~Goodfellow, and A.~Madry, ``On evaluating adversarial robustness,''
  \emph{arXiv preprint arXiv:1902.06705}, 2019.

\bibitem{gourdeau2019hardness}
P.~Gourdeau, V.~Kanade, M.~Kwiatkowska, and J.~Worrell, ``On the hardness of
  robust classification,'' in \emph{Advances in Neural Information Processing
  Systems}, 2019, pp. 7444--7453.

\bibitem{chen2017zoo}
P.-Y. Chen, H.~Zhang, Y.~Sharma, J.~Yi, and C.-J. Hsieh, ``Zoo: Zeroth order
  optimization based black-box attacks to deep neural networks without training
  substitute models,'' in \emph{Proceedings of the 10th ACM Workshop on
  Artificial Intelligence and Security}, 2017, pp. 15--26.

\bibitem{sharif2016accessorize}
M.~Sharif, S.~Bhagavatula, L.~Bauer, and M.~K. Reiter, ``Accessorize to a
  crime: Real and stealthy attacks on state-of-the-art face recognition,'' in
  \emph{Proceedings of the 2016 acm sigsac conference on computer and
  communications security}, 2016, pp. 1528--1540.

\bibitem{roth2019odds}
K.~Roth, Y.~Kilcher, and T.~Hofmann, ``The odds are odd: A statistical test for
  detecting adversarial examples,'' in \emph{International Conference on
  Machine Learning}, 2019, pp. 5498--5507.

\bibitem{pang2019mixup}
T.~Pang, K.~Xu, and J.~Zhu, ``Mixup inference: Better exploiting mixup to
  defend adversarial attacks,'' \emph{arXiv preprint arXiv:1909.11515}, 2019.

\bibitem{gilmer2019adversarial}
J.~Gilmer, N.~Ford, N.~Carlini, and E.~Cubuk, ``Adversarial examples are a
  natural consequence of test error in noise,'' in \emph{International
  Conference on Machine Learning}, 2019, pp. 2280--2289.

\bibitem{simonyan2014deep}
K.~Simonyan and A.~Zisserman, ``Very deep convolutional networks for
  large-scale image recognition,'' 2014.

\bibitem{rauber2017foolbox}
\BIBentryALTinterwordspacing
J.~Rauber, W.~Brendel, and M.~Bethge, ``Foolbox: A python toolbox to benchmark
  the robustness of machine learning models,'' \emph{arXiv preprint
  arXiv:1707.04131}, 2017. [Online]. Available:
  \url{http://arxiv.org/abs/1707.04131}
\BIBentrySTDinterwordspacing

\bibitem{ding2019advertorch}
G.~W. Ding, L.~Wang, and X.~Jin, ``{AdverTorch} v0.1: An adversarial robustness
  toolbox based on pytorch,'' \emph{arXiv preprint arXiv:1902.07623}, 2019.

\bibitem{art2018}
\BIBentryALTinterwordspacing
M.-I. Nicolae, M.~Sinn, M.~N. Tran, B.~Buesser, A.~Rawat, M.~Wistuba,
  V.~Zantedeschi, N.~Baracaldo, B.~Chen, H.~Ludwig, I.~Molloy, and B.~Edwards,
  ``Adversarial robustness toolbox v0.10.0,'' \emph{CoRR}, vol. 1807.01069,
  2018. [Online]. Available: \url{https://arxiv.org/pdf/1807.01069}
\BIBentrySTDinterwordspacing

\end{thebibliography}

\end{document}